# TLETA: Deep Transfer Learning and Integrated Cellular Knowledge for Estimated Time of Arrival Prediction

Hieu Tran, Son Nguyen, I-Ling Yen, Farokh Bastani
Department of Computer Science, The University of Texas at Dallas, USA
{trunghieu.tran, sonnguyen, ilyen, Farokh.Bastani}@utdallas.edu

*Abstract*—Vehicle arrival time prediction has been studied widely. With the emergence of IoT devices and deep learning techniques, estimated time of arrival (ETA) has become a critical component in intelligent transportation systems. Though many tools exist for ETA, ETA for special vehicles, such as ambulances, fire engines, etc., is still challenging due to the limited amount of traffic data for special vehicles. Existing works use one model for all types of vehicles, which can lead to low accuracy. To tackle this, as the first in the field, we propose a deep transfer learning framework TLETA for the driving time prediction. TLETA constructs cellular spatial-temporal knowledge grids for extracting driving patterns, combined with the road network structure embedding to build a deep neural network for ETA. TLETA contains transferable layers to support knowledge transfer between different categories of vehicles. Importantly, our transfer models only train the last layers to map the transferred knowledge, that reduces the training time significantly. The experimental studies show that our model predicts travel time with high accuracy and outperforms many state-of-the-art approaches.

## I. INTRODUCTION

The ETA problem (estimated time of arrival) has long been an important research direction in Intelligent Transportation Systems (ITS), especially after the rapid advances in IoT technologies [1]. ETA allows us to monitor the road traffic, effectively manage public transportations, and improve the quality of online ride-hailing and navigation platforms such as Google Map, Uber, Lyft, etc. [2] [3] [4]. Also, the rapidly increasing trajectory data collected from the Global Positioning System (GPS) have further promoted ETA studies. Some ETA solutions simply calculate the average travel time from a source to a destination based on historical GPS trajectory data [5] [6] [7]. Other techniques partition a trajectory into several segments and obtain different representations for each segment by some embedding methods for various learning models [3] [4] [8]. Recently, more studies have focused on a fine-grained data-driven approach that constructs models to learn spatial-temporal knowledge from small regions of the map where the route passes through [9] [10] [11] [12].

Although achieving positive results, these methods focus on dealing with problems such as GPS feature extraction and sparse routes. However, some environmental factors, such as weather, traffic-related events, etc., can consistently impact the driving time but have not been considered systematically for ETA. Some works mention the use of weather conditions for ETA but do not describe the specific information being used and how to extract the features [2] [4] [7]. Also, the road network structure, points of interest (POIs) distribution, etc. directly influence the traffic flow. A few research works consider POIs [9] [13] and road network structures [2] [12] [14], but they only preserve the local structure of the road network and ignore the global structure. Most importantly, existing works only consider the impact of individual environmental factors without considering the integrated impact. Integrated consideration of all common ETA impacting features may cause the problem of data sparseness and should be addressed correspondingly.

Due to different maneuvers on the road, different types of vehicles can have quite different driving times. Using one ETA prediction model for all vehicles can yield low accuracy for special vehicles such as ambulances, buses, police cars, etc. But building models for a specific vehicle type may suffer from the problem of scarce data or even no data. [15] showed that the amount of data for regular vehicle is around five times more than those for service vehicles in many cities. In Cincinnati, the data for service vehicles is only about 10% of all reported vehicle data [16]. Though some established methods, such as collaborative filtering [17], irregular convolution [18], training intensity modification [19], etc., can handle sparse data, they are not sufficiently effective to deal with extremely scarce data and the learning time can be very long.

On the other hand, transfer learning(TL) is a technique that efficiently utilizes knowledge of a learned model to solve another task in another domain. TL can significantly reduce the demand for resources, such as training data and training computation power, for a target domain by taking advantage of a pre-trained model from a large amount of source domain data [20]. Thus, leveraging TL techniques could address the scarce data problem better. Recent research works [9] [10] [21] [22] [23] [24] [25] proposed learning approaches to apply TL in ETA. However, these approaches use the entire pre-trained model from the source domain to either directly predict ETA for the target domain or re-train the entire model in the target domain, leading to lower accuracy and insufficient reduction of resource demands. Additionally, none of the existing works consider ETA discriminately based on different vehicle types.

In this paper, we propose a deep transfer learning framework (TLETA) for ETA. Our approach leverages the pretrained hidden layers from the source domain and only trains last layers in the target domain models, and thus, the training time and resource demands can be reduced significantly. Also, we consider several impactful environmental factors in an integrated way to improve ETA prediction accuracy. Some techniques in our approach are summarized in the following.

*Cellular spatial-temporal knowledge.* We collected data from different sources, including GPS trajectory data for regular and service vehicles [16]; static data including POIs, maps, holidays [16] [26]; and dynamic data such as weather conditions, traffic and social events [27] [28] to assist with ETA. For finer-grained mining of traffic patterns based on collected data, we partition the map into grid cells and construct different cellular spatial-temporal knowledge, including domain-specific and cross-domain knowledge. Furthermore, GPS trajectory data is deeply analyzed to compute the cellular speed knowledge. However, fine-grained learning can face sparse cellular knowledge even when we have sufficient data. Thus, we introduce an inner-domain data interpolation learning technique to handle the sparse data issue, which leverages the data-rich cells to generate the knowledge in data-sparse cells.

*Cellular learning model*. We propose a novel deep transfer learning framework for cellular spatial-temporal knowledge transferring among different vehicle types to solve the ETA problem efficiently and handle the scarce data problem for special vehicles. While inner-domain data interpolation learning is suitable for low-level sparsity in each domain to enrich the knowledge, the transfer learning approach can transfer knowledge from dense data domain to highly scarce data domain. First, grid cells are classified into different groups using a deep neural network based on the similarity of the average driving speed pattern. Then historical GPS trajectories are utilized to construct a directed road network graph encoded to low-dimensional representation by an autoencoder model for the local and global road structure preservation. The cellular classification output, road network embedding, and multiple cellular knowledge are used for the travel time prediction neural network model at cellular level. To transfer the learning knowledge among domains, the classifier model and travel time model contain the transferable layers trained in source domain and are applied for target models to improve the prediction performance. At the target domain models, only the last layers, SoftMax and fully connected customized layers for classifier and travel time model, respectively, will be trained and tuned that can boost the training process remarkably.

Given a GPS trajectory, we use a task-oriented algorithm that utilizes the cellular learning output for driving time estimation in real-time. The algorithm predicts the ETA based on the current time for the current cell and proceeds to the next cell with updated time till the destination is reached.

*Experimental study*. We demonstrate that our model can predict travel time with high accuracy for the target domain with a limited amount of data and outperforms the state-of-the-art approaches in terms of accuracy and training time.

The rest of this paper is organized as follows. Section II discussed related works. Section III introduces the problem specification. In Section IV, we propose our architecture framework and details of the models. The detailed experiment setup and results analysis are shown in Section V. Finally, Section VI concludes the work.

## II. RELATED WORK: ETA BY TRANSFER LEARNING

Transfer learning has recently been studied in ETA research. In [9], a spatial-temporal cross-domain neural network (STCNet) is introduced to capture the complex patterns hidden in cellular data from static data. STCNet learned the pattern diversity and similarity of cellular traffic of different city functional zones using a clustering algorithm and applied transferred knowledge between regions. A transfer learning method, RegionTrans [21], is proposed to facilitate deep spatial-temporal prediction in a data-scarce target city by transferring knowledge from a data-dense source city considering bike trajectory data and weather data to measure the crowd flow in vast cellular regions.

Some other studies focus on road segment learning. A clustering-based transfer learning model is proposed by Lin et al. [29], which extracts various spatial features in multiple levels and combines them with temporal features to support this transfer learning scenario. Similarly, [22] described a Feature-based Transfer Learning (FBTL) to transfer the spatial-temporal road segment speed learning between regions using GPS and weather data. Recently, Huang et al. proposed a transfer learning approach with a graph neural network (TEEPEE) that uses graph clustering to divide the traffic network map into multiple sub-graphs [23]. Graph clustering can capture more spatial information in the transfer process. Lately, Mallick T. developed TL-DCRNN, a transfer learning approach, where a single model trained on a highway network can be used to forecast traffic on unseen highway networks [24]. The above works mainly focus on GPS data and pay less attention to other factors.

Transfer learning has also been applied among regions. In [25], a deep learning method is introduced which utilizes OSM data to build spatial-temporal speed knowledge for continuous traffic speed prediction on a road segment. [30] integrated the OSM data with live traffic data from Baidu map to label the traffic condition in various parts of Shanghai. Transfer learning is applied from one district region to another. In [10] an open-source pre-trained deep learning network is used to extract the relevant features from the traffic imagery data to automatically identify different network traffic statuses. These approaches transfer the entire model directly from the source domain to the target domain, which may result in low prediction accuracy.

In general, there is no transfer learning work for ETA for knowledge transfer among different types of vehicles.

## III. PROBLEM SPECIFICATION

**Definition 1**. *Cell*. Consider a grid map $\mathcal{M}$ of a region. We partition $\mathcal{M}$ into $I \times J$ grid cells where each cell represents a spatial region of $\varphi° \times \varphi°$ (latitude × longitude), and we call $\varphi°$ the splitting factor. All the cells in the region form a set $C = \{c^{h,w} | 1 \leq h \leq I, 1 \leq w \leq J\}$, where $c^{h,w}$ denotes the cell of the $h$-th row and the $w$-th column of $\mathcal{M}$.

Using grid cells, instead of road segments and links, to represent the road network can offer finer-granularity and more consistency in travel speed estimations.

**Definition 2**. *GPS trajectory*. A GPS trajectory $Gt$ is a sequence of $|Gt|$ GPS points $p_i$, i.e., $Gt=\langle p_1, p_2, p_3, ..., p_{|Gt|}\rangle$, where $p_i=(\alpha_i, \beta_i, t_i)$, including longitude $\alpha_i$, latitude $\beta_i$, and timestamp $t_i$.

**Definition 3**. *Spatial-temporal knowledge.* We add time to the geographical cell map $\mathcal{M}$ and construct the 3D spatial-temporal $I \times J \times T$ grid to finely describe the traffic patterns, where $T$ is the number of time intervals in one day. At time $t \in T$, we have an image $\mathcal{I}$ of map $\mathcal{M}$ with size $I \times J$ to represent the set of traffic dependency factors $td$,

$$\mathcal{I}_{td,t} = \{r_{td,t}^{h,w} | 1 \leq h \leq I, 1 \leq w \leq J\} \in \mathbb{R}^{I \times J}$$

where $r_{td,t}^{h,w}$ is the traffic dependency factors for cell $(h,w)$ at time $t$, including weather, social events, traffic events, etc.

**Problem 1**. *Travel Time Prediction.* Given vehicle $V$ and corresponding GPS trajectory $Gt$ from the source $p_{src}=p_1$ to the destination location $p_{des}=p_{|Gt|}$, we aim to find a mapping $f: Gt \rightarrow F$ that takes the GPS trajectories $Gt$ as inputs and predicts the travel time between $p_{src}$ and $p_{dest}$ via trajectory $Gt$ using traffic dependence factors. Assume that trajectories are selected by users or generated by planning applications.

**Problem 2**. *Transfer learning for cross-vehicle ETA prediction.* Since each type of vehicle has different maneuvers on the road, their travel time may be different. Some vehicles may have a lot of historical data for training, but others may not, which can cause difficulties in ETA prediction. Given the *dense* GPS trajectory data for vehicle $V$ in map $\mathcal{M}$ and the *scarce* GPS data for vehicles $V'$, we aim to learn the function $f'$ to predict the driving time for vehicles $V'$ using knowledge transferred from vehicles $V$.

## IV. METHODOLOGY AND FRAMEWORK

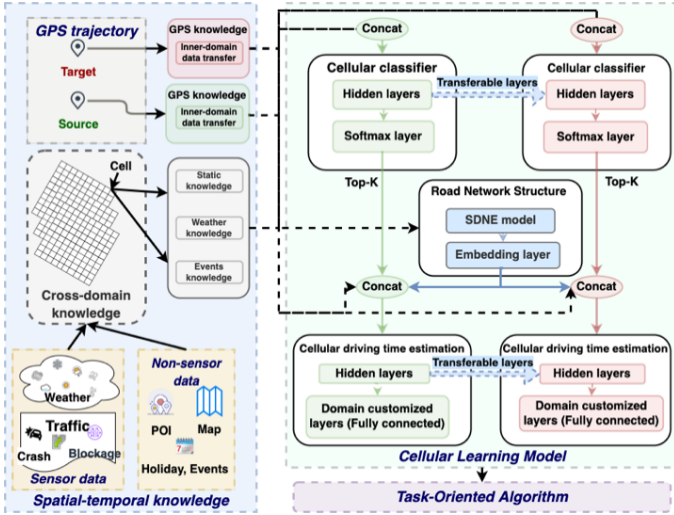

Fig. 1. TLETA: Deep Transfer Learning and Integrated Cellular Data for Estimated Time of Arrival Prediction Architecture

Fig. 1 shows the architecture for TLETA, a deep transfer learning framework for vehicle travel time prediction. TLETA contains 3 major modules: cellular spatial-temporal knowledge extraction (CKE), cellular learning (CL), and task-oriented prediction (TOP) modules. CKE integrates raw data collected from various sources into cellular spatial-temporal knowledge. The CL module includes a traffic pattern classifier, a road network embedding scheme, and an ETA algorithm that learns the cellular traffic and makes cellular level travel time predictions. The task-oriented prediction module uses the learned model to continuously predict ETA for the given trajectory.

### A. Cellular Spatial-Temporal Knowledge

We divide the cellular knowledge into two categories: cross-domain and domain-specific. The cross-domain knowledge describes the environment and is applicable to all vehicles. The domain-specific knowledge is extracted from the GPS data for a specific vehicle type. We define three categories of cross-domain knowledge, followed by domain-specific knowledge.

**Static knowledge**. Intuitively, traffic condition is affected by the density of the local road network and the POIs in the area. For example, more traffic lights may imply slower traffic flow. More office buildings nearby a region implies a higher traffic congestion during rush hours. Since POI is static and rarely changes over time, we drop the time dimension and use a 2D spatial knowledge matrix for its representation.

$$\mathcal{J}_{POI} = \begin{bmatrix} r_{POI}^{1,1} & r_{POI}^{1,2} & \cdots & r_{POI}^{1,J} \\ r_{POI}^{2,1} & r_{POI}^{2,2} & \cdots & r_{POI}^{2,J} \\ \vdots & \vdots & \ddots & \vdots \\ r_{POI}^{I,1} & r_{POI}^{I,2} & \cdots & r_{POI}^{I,J} \end{bmatrix}$$

where $r_{POI}^{h,w}$ is the POI density of cell $(h,w)$, and $\mathcal{J}_{POI} \in \mathbb{R}^{I \times J}$. Besides, we also consider whether the travel day is a weekend or holiday time as $r_{date}=\{weekend, holiday | weekend, holiday \in \{0,1\}\}$.

**Weather knowledge**. Several weather factors, including rain level, snow level, hail level, are used to categorize traffic. The matrix of the weather knowledge is defined as follows.

$$\mathcal{J}_{weather,t} = \begin{bmatrix} r_{weather,t}^{1,1} & r_{weather,t}^{1,2} & \cdots & r_{weather,t}^{1,J} \\ r_{weather,t}^{2,1} & r_{weather,t}^{2,2} & \cdots & r_{weather,t}^{2,J} \\ \vdots & \vdots & \ddots & \vdots \\ r_{weather,t}^{I,1} & r_{weather,t}^{I,2} & \cdots & r_{weather,t}^{I,J} \end{bmatrix}$$

where $r_{weather,t}^{h,w}$ is the weather knowledge for $(h,w)$ at time $t$.

**Event knowledge.** Traffic and social events also play a vital role in travel time estimation. A blockage or crash event on the freeway can severely delay the traffic flow. The social events also impact the traffic in a region. We express the event knowledge matrix as follows.

$$\mathcal{J}_{event,t} = \begin{bmatrix} r_{event,t}^{1,1} & r_{event,t}^{1,2} & \cdots & r_{event,t}^{1,J} \\ r_{event,t}^{2,1} & r_{event,t}^{2,2} & \cdots & r_{event,t}^{2,J} \\ \vdots & \vdots & \ddots & \vdots \\ r_{event,t}^{I,1} & r_{event,t}^{I,2} & \cdots & r_{event,t}^{I,J} \end{bmatrix}$$

where $r_{event,t}^{h,w}$ denotes the number of traffic and important social events at time $t$ in cell $(h,w)$.

**Domain-specific knowledge (GPS knowledge)**. Due to different maneuvers, different types of vehicles have different driving time characteristics. We construct cellular spatial-temporal knowledge for each vehicle domain. Also, for fine-grained understanding of traffic patterns at the cellular level, for each vehicle type, we derive the travel speed in each cell from the *GPS trajectory data*. (Note that the trajectory data for different days are grouped together, so time intervals should be considered as circular, and before the first interval is the $T$-th.)

Consider two consecutive data points, $p_1 = \langle \alpha_1, \beta_1, t_1 \rangle$ and $p_2 = \langle \alpha_2, \beta_2, t_2 \rangle$ on a GPS trajectory, where $\alpha_1$ and $\alpha_2$ are the latitudes, $\beta_1$ and $\beta_2$ are the longitudes, and timestamps $t_1 \leq t_2$. Let $\Delta\alpha_{1,2}$ and $\Delta\beta_{1,2}$ denote the differences between $\alpha_1$ and $\alpha_2$ and between $\beta_1$ and $\beta_2$, respectively. Let $R$ be the radius of the Earth. The speed between $p_1$ and $p_2$ is $s_{1,2} = d_{1,2}/(t_2 - t_1)$, where the distance $d_{1,2}$ between $p_1$ and $p_2$ can be derived as:

$$d_{1,2} = 2R * atan2(\sqrt{sin^2(\Delta\alpha_{1,2}/2) + cos\alpha_1 * cos\alpha_2 * sin^2(\Delta\beta_{1,2}/2)}, \sqrt{1 - sin^2(\Delta\alpha_{1,2}/2) - cos\alpha_1 * cos\alpha_2 * sin^2(\Delta\beta_{1,2}/2)})$$

If two consecutive GPS data points $p_1$ and $p_2$ on a trajectory falls in the same cell, we compute its speed and use it directly as a sample for the cell. Otherwise, we identify the set of cells $SC_{1,2}$ that the segment between $p_1$ and $p_2$ crosses. To derive $SC_{1,2}$, we find the midpoint between $p_1$ and $p_2$ to divide the segment into two. For each of the new segments, we further divide it in the same way till every segment falls in a single cell in $SC_{1,2}$. Each segment is then used as a sample for deriving the travel speed in its corresponding cell. We use the same speed $s_{1,2}$ as the speed of all segments between $p_1$ and $p_2$.

The midpoint $p_{mid} = \langle \alpha_{mid}, \beta_{mid} \rangle$ between $p_1$ and $p_2$ can be derived as:

$$\alpha_{mid} = atan2(sin\alpha_1 + sin\alpha_2, \sqrt{(cos\alpha_1 + cos\alpha_2 * cos\Delta\beta_{1,2})^2 + (cos\alpha_2 * sin\Delta\beta_{1,2})^2})$$
$$\beta_{mid} = \beta_1 + atan2(cos\alpha_2 * sin\Delta\beta_{1,2}, cos\alpha_1 + cos\alpha_2 * cos\Delta\beta_{1,2})$$

Besides the speed, the bearing is also an essential factor affecting the driving time since traffic flow in two ways of the road can be different even if it is for the same time in the same cell. The bearing of travel from $p_1$ to $p_2$ is:

$$bear_{1,2} = atan(sin\Delta\beta_{1,2} * cos\alpha_2, cos\alpha_1 * sin\alpha_2 - sin\alpha_1 * cos\alpha_2 * cos\Delta\beta_{1,2})$$

The bearing value changes frequently even when the vehicle moves in the same direction. To avoid considering minor changes, we group the bearing into eight directions: $dir_{1,2} = direction(bear_{1,2}) \in RD$, and $RD = \{N, NE, E, SE, S, SW, W, NW\}$.

The knowledge matrix based on GPS trajectories is defined as follows:

$$J_{GPS,t} = \begin{bmatrix} r_{GPS,t}^{1,1} & r_{GPS,t}^{1,2} & \cdots & r_{GPS,t}^{1,J} \\ r_{GPS,t}^{2,1} & r_{GPS,t}^{2,2} & \cdots & r_{GPS,t}^{2,J} \\ \vdots & \vdots & \ddots & \vdots \\ r_{GPS,t}^{I,1} & r_{GPS,t}^{I,2} & \cdots & r_{GPS,t}^{I,J} \end{bmatrix}$$

where $r_{GPS,t}^{h,w}$ represents the information at cell $(h, w)$ at time $t$. Due to the different traffic bearings, $r_{GPS,t}^{h,w} = \{\langle dir_i^{h,w}, s_i^{h,w} \rangle\}$ includes $ns$ tuples ($ns = |r_{GPS,t}^{h,w}|$), each represents the cellular speed $s_i^{h,w}$ corresponding to a direction $dir_i^{h,w}$.

*Inner-domain data interpolation learning.* Some cells may not have GPS trajectory data at some time intervals. To supplement the missing knowledge, another GPS knowledge matrix $J_{GPS,t}^{T^a}$ is constructed based on the historical trajectories. At cell $(h,w)$ at time $t$, GPS knowledge for speed can be computed by the average of $T^a$ ($T^a \leq T/2$) time intervals before and after $t$ in the same direction.

We construct an inner-domain data learning model $MI$ to estimate the GPS knowledge for missing GPS data using $J_{GPS,t}$ and $J_{GPS,t}^*$, where $J_{GPS,t}^* = \{J_{GPS,t}^{T^a} | 2 \leq T^a \leq T/2\}$. The model can learn the traffic pattern in data-rich cells and use it to estimate the traffic pattern in data-sparse cells. For simplicity, the one-layer supervised model $MI$ receives input $r_{GPS,t}^{h,w} \oplus r_{GPS,t}^{h,w\ *}$ for cell $(h, w)$ and outputs the speed prediction, where $\oplus$ is the concatenation operation. We construct the new mixed knowledge matrix based on the output of each cell, i.e.,

$$r_{GPS,t}^{h,w} = f_{MI}(r_{GPS,t}^{h,w\ *}, r_{GPS,t}^{h,w})$$

where $f_{MI}$ denotes the activation function of $MI$.

**Cellular spatial-temporal knowledge**. Overall, at time $t$ in cell $(h, w)$, we have cross-domain and domain-specific cellular knowledge, which can be represented as:

$$r_t^{h,w} = \{r_{date}, r_{POI}^{h,w}, r_{weather,t}^{h,w}, r_{event,t}^{h,w}, r_{GPS,t}^{h,w}\}$$

**Embedding**. Embedding methods have been widely used to transform data in a high-dimension space into a low-dimension embedding vector that can be used to represent the hidden information of the object. Both numerical and categorical knowledge will be embedded into low-dimension vectors for learning models. Numerical data values will be normalized into the range [0,1], and the categorical values will be encoded into 1-hot $m$-dimension binary vector.

### B. Cellular Learning Model

The cellular learning models in TLETA contain three components: cellular classifier model, road network structure, and cellular driving time estimation.

#### 1) Cellular classifier model

As a common knowledge, the traffic of a cell at a particular time of each day often has the same pattern if there is no special or unexpected weather, events, etc. Also, two cells can have similar traffic patterns at the same or different times. Thus, instead of simply using time of the day for grouping data, we classify cells with similar traffic patterns into the same groups to achieve better learning results. Specifically, we classify the spatial-temporal cells into $\mathcal{N}$ different categories of traffic levels based on average speed and other cellular knowledge. A neural-net classification model that contains $L_1$ hidden layers and a SoftMax layer $\sigma$ is used to classify cells into traffic levels. The input of the model is defined as follows.

$$\mathbb{I} = \{r_t^{h,w}, \overline{r_{GPS,t}^{h,w}}\}$$

$$\overline{r_{GPS,t}^{h,w}} = \frac{\mathcal{N}}{Max_{1 \leq i \leq ns}(r_{GPS,t}^{h,w} \cdot s_i^{h,w})} \frac{1}{ns} \sum_{i=1}^{ns} r_{GPS,t}^{h,w} \cdot s_i^{h,w}$$

where $\overline{r_{GPS,t}^{h,w}}$ is the average speed level of all GPS knowledge for cell $(h, w)$ at the time $t$. The model output is the probability from SoftMax layer $\sigma$ as the class membership confidence, i.e.,

$$\sigma(z)_i = \frac{e^{z_i}}{\sum_{j=1}^{N} e^{z_j}}$$

where $z$ is the embedding vector input to layer $\sigma$. Top-k classes $\mathcal{K}$ from the SoftMax layer are chosen as embedding features for the driving time estimation model. Hyperparameter $k$ defines the degree of approximation by indicating the number of classes used for ETA prediction.

$$\mathcal{K} = argmax_{\sigma' \subset \sigma, |\sigma'|=k} \sum_{p \in \sigma'} p$$

### 2) Road network structure

The road network structure can impact the travel time too. For example, driving on one-way or two-way roads or around an intersection can lead to different driving times for the same vehicle. We construct a road network graph to extract the road network knowledge via embedding layers.

**Definition 4**. *Upstream cells*. Given a grid cell $c$, upstream cells $UC_c$ of $c$ is the set of cells that can reach $c$ in one step. For example, assume two GPS trajectories passing by the grid cells: $c_1 \to c_3 \to c_2 \to c_4$ and $c_5 \to c_6 \to c_2 \to c_7$. The upstream cells of $c_2$ are $UC_{c_2} = \{c_3, c_6\}$. The upstream cells of a cell can be determined from GPS trajectory and OSM data.

**Definition 5**. *Road network*. Road network structure $\mathcal{G}(\mathcal{V}, \mathcal{E})$ describes a directed graph constructed based on upstream cells. The vertex set $\mathcal{V}$ represents the cells ($\mathcal{V} = I \times J$), and the edge set $\mathcal{E}$ denotes the connected relationship among cells. Let $S_{adj}$ be the adjacent matrix of $\mathcal{G}(\mathcal{V}, \mathcal{E})$, where $s_{i,j}$ denotes whether $c_i$ and $c_j$ are connected, i.e.

$$s_{i,j} = \begin{cases} 1, & if\ c_i \in UC_{c_j} \\ 0, & otherwise \end{cases}$$

Network embedding is essential for integrating the road network structure into cellular knowledge learning. When deriving similarity of the cells, it is necessary to consider both the local and global network structures. Correspondingly, we define the local network structure as the first-order proximity and the global network structure as the second-order proximity.

**Definition 6**. *First-order proximity* in the network is the local pairwise proximity between two cells. For any pair of cells $c_i$ and $c_j$, the first-order proximity exists if $s_{i,j} = 1$, otherwise the local proximity is 0.

**Definition 7**. *Second-order proximity* between two cells represents the similarity of their neighborhood structure. Let $NS_i = \{s_{i,1}, s_{i,2}, \ldots, s_{i,|\mathcal{V}|}\}$ denote the first-order proximity of $c_i$ with all other cells describing the neighborhood structure. The second-order proximity between $c_i$ and $c_j$ is defined by the similarity of $NS_i$ and $NS_j$.

Intuitively, the first-order proximity assumes that two cells in a local region are similar if they are connected in the road network. Similarity of the second-order proximity requires two cells to share many common neighbors, no matter whether they are directly connected. We use semi-supervised deep model SDNE [31], an autoencoder for graph embedding, to capture the low-dimensional representation of cells in the road network. The encoder in SDNE maps the input data into low-dimensional vector $\omega$, while the decoder maps $\omega$ back to the original input representation. Given the neighborhood structure $NS_i$ of cell $c_i$ as input, the hidden representation for each layer is defined as follows:

$$y_i^{(1)} = \sigma(W_{NS}^{(1)} * NS_i + b^{(1)})$$
$$y_i^{(k)} = \sigma(W_{NS}^{(k)} * y_i^{(k-1)} + b^{(k)}), k = 2..K; \omega = y^K$$

where $W_{NS}^{(k)}$ denotes the learnable parameter matrix, and $b^{(k)}$ indicates the bias terms for the sigmoid activation function $\sigma$ with $\sigma(x) = 1/(1 + \exp(-x))$. Since it is essential to preserve the local road network structure, we define the loss function for the first-order proximity as follows:

$$\mathcal{L}_1 = \sum_{i,j=1}^{\mathcal{V}} s_{i,j} \left\| y_i^{(K)} - y_j^{(K)} \right\|_2^2$$

where $\|\cdot\|_2^2$ denotes the $\mathcal{L}2$-norm.

The decoder returns $\widehat{NS_i}$ of cell $c_i$. The goal of autoencoder is to minimize the reconstruction loss of the input and output data. Since the input is the neighborhood structure of each cell, the reconstruction process will tend to make cells with similar neighborhood structures to have similar $\omega$. These cells will be mapped close to each other in the $\omega$ space. The loss function of second-order proximity can be computed as:

$$\mathcal{L}_2 = \sum_{i=1}^{\mathcal{V}} \left\| (\widehat{NS_i} - NS_i) \odot b_i \right\|_2^2 = \left\| (\widehat{NS} - NS) \odot b \right\|_F^2$$

where $\odot$ means the Hadamard product, $b_i = \{b_{i,j}\}_{j=1}^{\mathcal{V}}$. If $s_{i,j} = 0$, $b_{i,j} = 1$, otherwise $b_{i,j} = \xi > 0$. Concatenating the loss function from the first-order and the second-order proximities, we obtain the final loss function for the road network learning model as follows:

$$\mathcal{L}_{road} = \gamma \mathcal{L}_1 + \mathcal{L}_2 + \frac{1}{2} \sum_{k=1}^{K} \left( \left\| W_{NS}^{(k)} \right\|_F^2 + \left\| \widehat{W}_{NS}^{(k)} \right\|_F^2 \right)$$

where $\gamma$ denotes the hyperparameter and $\widehat{W}_{NS}^{(k)}$ are parameter matrices in the decoder process. The third term is an $\mathcal{L}2$-norm regularization term to prevent over-fitting. After minimizing the loss function, we obtain the vector $\omega$ for each cell that preserves the local and global structure of the road network.

### 3) Cellular driving time estimation model

We integrate road network structure embedding, top-k classes $\mathcal{K}$ from the SoftMax layer $\sigma$, and domain-specific and cross-domain knowledge into one prediction model. Since we only consider $\mathcal{K}$ classes, we normalize $\sigma$ to $\sigma^{norm}$ as follows:

$$\sigma(z)_i^{norm} = \begin{cases} \sigma(z)_i & if\ \sigma(z)_i \in \mathcal{K} \\ 0, & otherwise \end{cases}$$

$\sigma^{norm}$, road network structure representation $\omega$, and all cellular knowledge discussed above are used as the input for the travel time learning model, i.e.,

$$\mathbb{I} = r_t^{h,w} \oplus \sigma^{norm} \oplus \omega$$

We construct a dense neural network with $L_2$ hidden layers followed by a domain customized layer $L_{dc}$, a fully connected layer. Each layer contains an activation function $f_l(\cdot)$. The first layer receives input $\mathbb{I}$, the $k$-th layer receives the computed features from the previous layer as follows:

$$y_0 = \mathbb{I};\ y_1 = f_1(y_0);\ y_k = f_k(y_{k-1})$$

Output from the last layer $L_{dc}$, $\hat{Y} = y_{L_{dc}} \in \mathbb{R}^{I \times J \times T}$, is the estimated cellular travel time. The objective function $\mathcal{L}(\Theta)$ is to minimize the error between the forecast value and the ground truth, where

$$\hat{Y} = y_{L_{dc}} = f_{L_{dc}}(y_{L_2}); \quad \mathcal{L}(\Theta) = argmin_\theta \|Y - \hat{Y}\|$$

where $\Theta$ denotes the model parameters that can be trained through optimization techniques.

*4) Transfer learning among different vehicle domains*

Problem 2 considers transfer learning of the cellular traffic knowledge between vehicle domains. We construct hidden layers to perform knowledge transfer between domains. In the neural-net cellular classifier, all hidden layers are transferable except for the SoftMax layer, a domain customized layer. The classifier model is constructed in the source domain as $M_1^S = L_1 \oplus \sigma^S$. The target domain's classifier model is built based on the transferable learning layers of the source domain as $M_1^T = L_1 \oplus \sigma^T$. The SoftMax layer is characterized for each domain, which is non-transferable. In the cellular driving time estimation model, like classifier models, the learning model can be expressed as $M_2^S = L_2 \oplus L_{dc}^S$ in the source domain and $M_2^T = L_2 \oplus L_{dc}^T$ in the target domain. Note that the last customized layer is fully connected and non-transferrable.

Transferable layers $L_1$ and $L_2$ are trained in the source domain with a large amount of data that are then used for training in the target domain. The learning model of the target domain is initialized with $L_1$ and $L_2$ with fixed parameters. Only the SoftMax and fully connected customized layers will be trained and tuned. Thus, it can boost the training process and improve accuracy for scarce data cases.

### C. Task-oriented algorithm

**Algorithm 1: Task-oriented algorithm**

**Input**: Source $src$ and destination $des$ location, GPS trajectory $gps$, timestamp $t$.
**Output**: Driving time estimation
1: **begin**
2: $curr = src$ ; $\mathbb{C}$ = convertGPStoCells($gps$)
3: **while not** reachDestination($curr$, $des$)
4:     $knowledge$ = collectKnowledge($t$, $curr$)
5:     $cell\_t$ = getDrivingTime($\mathbb{C}$ , $curr$, $t$, $knowledge$)
6:     $curr$ = nextCell($\mathbb{C}$, $curr$)
7:     updateResult($result$, $cell\_t$)
8:     updateFutureTime($t$, $cell\_t$)
9: **end while**
10: **end**

We introduce a task-oriented algorithm, for ETA prediction (Algorithm 1). Given a GPS route, including source and destination locations, we aim to calculate the travel time for the route. First, the algorithm converts GPS data points into corresponding cellular data $\mathbb{C}$. Then, cellular knowledge data are gathered for the cells on the route. After computing the driving time of the source cell based on the cellular learning model, a new timestamp is considered for the next position on the route. The process continues recursively and ends when we reach the destination location.

## V. EXPERIMENTAL STUDIES

We evaluate our TLETA framework and compare it with several state-of-the-art methods using a real-world, large-scale dataset. Two experiments are conducted to evaluate the two major designs in TLETA: (1) the cellular knowledge learning model and (2) the transfer learning approach.

**Data collection**. We collected various real-world data from different sources for the City of Cincinnati for the entire year of 2018. The main data source is the GPS trajectory data [16], which specifies the vehicle type for each trajectory. Accordingly, we split the dataset into the regular vehicle domain (RV) and the service vehicle domain (SV). For each domain, we further divided the datasets based on their regions, including the urban (RVU of 1.2M and SVU of 1M data points) and suburban (RVS of 1M and SVS of 175K data points). Among them, SVS data is scarce and transfer learning could help most. For each dataset, we use 70% data for training, 10% for validation, and 20% for testing.

For static road network data, we collected POI points from OSM [26] for the Cincinnati area, including roads, trails, cafes, stores, restaurants, etc. The traffic event data are collected from [16], including crash and blockage events. The social events, including music events, sport games, conferences, etc., are obtained by crawling the Web resources related to the city, including social networks and event websites. The weather data is retrieved from [27] [28]. Holidays, weekends, and other special days are directly identified from the calendar.

**Experimentation setup**. Our experimental evaluation is deployed in TensorFlow 2.3.0 and runs on a workstation with Intel Xeon W-2145 CPU 3.7 GHz, 32GB DDR, GPU NVIDIA Quadro P5000 32GB. We used the same set of hyperparameter values for all the training. The transferable hidden layers consist of 3 layers for the cellular driving time model and one for the classifier. The unit number is adjustable in the range [16,1024], and each unit has a ReLU activation function. We use Adam optimizer [32] with a learning rate of 0.001. To prevent overfitting, dropout layers are used with the dropout rate of 0.1. The batch size is set to 32. Top-k classes from the SoftMax layer is set to k=5. In the road network structure component, the layer number in the encoder and decoder is 2. Splitting factor $\varphi°$ is set to $0.001°$.

**Performance metrics**. The accuracy metrics we consider include Mean Absolute Percentage Error (MAPE) and Root Mean Square Error (RMSE).

$$MAPE(y, \hat{y}) = \frac{100\%}{N} \sum_{i=1}^{N} \left| \frac{y^{(i)} - \hat{y}^{(i)}}{y^{(i)}} \right|;$$
$$RMSE(y, \hat{y}) = \sqrt{\frac{1}{N} \sum_{i=1}^{N} (y^{(i)} - \hat{y}^{(i)})^2}$$

where $y^{(i)}$ is the ground truth, $\hat{y}^{(i)}$ represents the predicted value, and $N$ is the number of instances in the test set.

### A. Evaluation of Our Cellular Knowledge Model

First, we evaluate our integrated cellular knowledge model for ETA and study the impact of each category of knowledge. The categories of cellular knowledge in our learning model in-clude GPS(G), Static(S), weather(W), events(E), and road network structure(R). We drop some categories of knowledge and study the impacts. We also compare our ETA model with the state-of-the-art methods that use less knowledge, including [11] which uses GPS and road network structure knowledge with the LSTM model, [18] which uses weather and events knowledge with irregular convolutional residual LSTM model, [6] which only uses GPS data in a similarity-based model, and [4] which uses weather and road network structure knowledge with a graph attention network model. Table I shows the comparison results.

TABLE I. Cellular knowledge model for ETA

| Approach | Knowledge | | | | | Metrics | |
|---|---|---|---|---|---|---|---|
| | G | S | W | E | R | MAPE | RMSE |
| [11] | x | | | | x | 19.57% | 76s |
| [18] | x | | x | x | | 18.44% | 74s |
| [4] | x | | x | | x | 17.92% | 71s |
| [6] | x | | | | | 25.32% | 102s |
| Reduced knowledge categories | x | | | | | 23.78% | 86s |
| | x | x | | | | 19.64% | 72s |
| | x | | x | | | 21.44% | 81s |
| | x | | | x | | 20.85% | 80s |
| | x | | | | x | 18.23% | 70s |
| | x | | x | x | | 18.20% | 70s |
| | x | | x | | x | 16.88% | 66s |
| | x | | x | x | x | 14.35% | 58s |
| | x | x | | x | x | 13.66% | 53s |
| | x | x | x | | x | 14.02% | 56s |
| | x | x | x | x | | 15.71% | 62s |
| | **x** | **x** | **x** | **x** | **x** | **12.37%** | **36s** |

As can be seen, integrated knowledge model achieved the best prediction result. Road network structure knowledge contributes more toward ETA prediction. This may be the impact from special road structures such as intersections, 2-way roads, etc. Also, our approach obtains better results than other ones which lacks complete understanding of relevant knowledge.

### B. Evaluation of Transfer Learning for ETA

We compare our approach with several state-of-the-art ETA works using transfer learning (TL), including STCNet [9], RegionTrans [21], TL-DCRNN [24], FBTL [22], and TEEPEE [23]. To study the improvement due to TL, we also consider a non-transfer model (TLETA that only uses target domain data).

TABLE II. Transfer learning performance comparison of TLETA and other traffic forecasting methods

| Dataset | Urban | | | Suburban | | |
|---|---|---|---|---|---|---|
| Approach | MAPE | RMSE | Time | MAPE | RMSE | Time |
| STCNet | 17.65% | 62s | 104m | 18.01% | 76s | 91m |
| RegionTrans | 16.23% | 56s | 110m | 17.88% | 67s | 93m |
| TL-DCRNN | 16.98% | 57s | 72m | 17.21% | 65s | 56m |
| FBTL | 23.29% | 81s | 122m | 25.78% | 105s | 99m |
| TEEPEE | 20.96% | 70s | 88m | 22.43% | 85s | 72m |
| Non-transfer | 14.78% | 55s | 49m | 18.51% | 74s | 39m |
| **TLETA** | **10.54%** | **34s** | **31m** | **11.81%** | **38s** | **29m** |

As shown in Table II, our model outperforms other approaches in MAPE, RMSE, and training time. The reasons include (1) they only uses a subset of knowledge categories, and (2) their models utilize complicated designs, such as convolutional neural networks and long/short-term memory models and, hence, require more training time. Since TLETA only trains the SoftMax and the fully connected customized layers, it boosts the training process significantly.

When the target domain has a considerable amount of data, we observe that TLETA still improves the prediction accuracy and can benefit from fast learning. In the case of scarce data (SVS), the prediction accuracy improves 36% for the transfer approach compared to the non-transfer one.

To understand the impact of the classifier component in TLETA, we compare our model with a version without the classifier (results in Table III). From the results, we can see that the absence of classifier leads to lower accuracy. This makes sense since the classifier groups the cells with similar traffic patterns together, which can improve learning effectiveness.

TABLE III. Component impact on transfer learning performance

| Dataset | Urban | | Suburban | |
|---|---|---|---|---|
| Version | MAPE | RMSE | MAPE | RMSE |
| W/o classifier | 19.75% | 74s | 21.54% | 74s |
| TLETA | 10.54% | 34s | 11.81% | 38s |

### C. Parameter Sensitivity

We analyze the impact of different time intervals (Fig. 2(a)), travel distances (Fig. 2(b)), numbers of events (Fig. 2(c)) on TLETA prediction accuracy, considering both the TL and the non-transfer model. As shown, MAPE peaks at rush hours, around 7-8 am and 6 pm. This is likely due to the unpredictability with busier and less stable traffic. Also, TLETA performs the best in a medium distance from 5-20 miles. Longer distance could incur a higher cumulative error. For shorter trips, the prediction within the beginning and ending cells may contribute more uncertainties. Moreover, with fewer events, TLETA yields better prediction accuracy. This could be due to the different impacts by different events and simple event counting does not provide sufficient knowledge. But as shown in Table I, even the less precise knowledge helps improve accuracy.

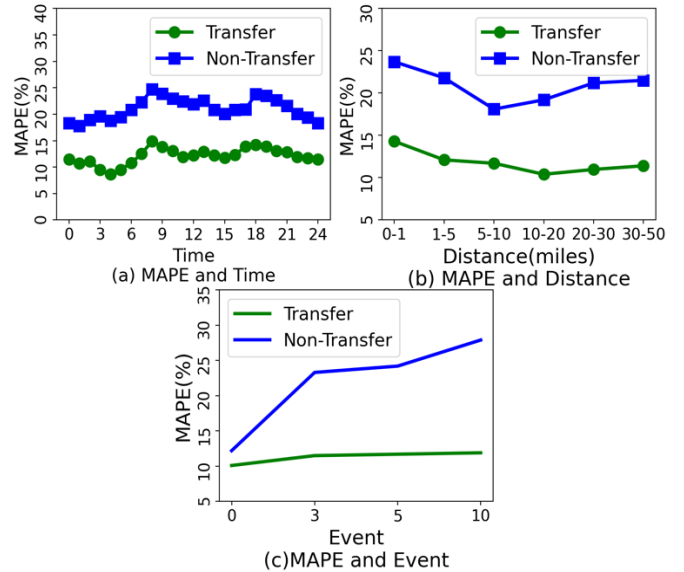

Fig. 2. MAPE sensitivity analysis

To learn the best parameters for TL, we measure the MAPE convergence based on the number of epochs and number of transferable units (Fig. 3). From Fig. 3(a), we can see that the learning curve of TL converges after around 10-15 epochs. Also, TL is 50% faster than the non-TL approach. From Fig. 3(b), we can see that the best choice for the number of transferable units should be in [256,512]. Transferring more units might not improve the accuracy, and it requires more resources for a heavier neural network model.

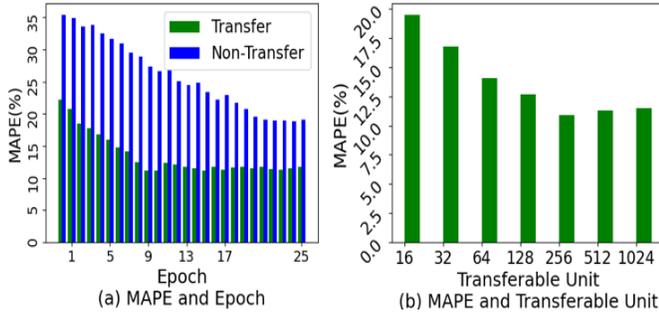

Fig. 3. Learning parameter analysis

## VI. CONCLUSION

This paper presented a novel deep transfer learning model TLETA for travel time prediction in different vehicle domains by leveraging a comprehensive set of cellular spatial-temporal knowledge and several learning mechanisms. Our learning mechanisms include a cellular classifier which classifies cells based on cellular spatial-temporal knowledge, a sparse data handler to derive missing knowledge based on the surrounding data points, and the travel time predictor which leverages the cellular knowledge and the classification results to make real-time ETA predictions. The ETA predictor is designed with transferable layers to transfer the spatial-temporal knowledge between domains of different vehicle types. Our approach outperforms the state-of-the-art ETA approaches in terms of accuracy and training time. Also, transfer learning made our approach highly effective even when the number of data points in certain domains is scarce.